\documentclass[10pt,twocolumn,letterpaper]{article}
\usepackage{iccv}
\usepackage{times}
\usepackage{epsfig}
\usepackage{graphicx}
\usepackage{amsmath}
\usepackage{amssymb}
\usepackage{amsmath}
\usepackage{galois}
\usepackage{caption}
\usepackage{subfigure}

\newcommand{\tabincell}[2]{
\begin{tabular}{@{}#1@{}}#2\end{tabular}
}

\usepackage[pagebackref=true,breaklinks=true,letterpaper=true,colorlinks,bookmarks=false]{hyperref}

 \iccvfinalcopy 

\def\iccvPaperID{747} 

\ificcvfinal\fi
\begin{document}

\title{Learning Deep Neural Networks for Vehicle Re-ID with \\Visual-spatio-temporal Path Proposals}

\author{
Yantao Shen$^{1}$ \quad 
Tong Xiao$^{1}$ \quad 
Hongsheng Li$^{1,}$\footnotemark[1] \quad 
Shuai Yi$^{2}$ \quad 
Xiaogang Wang$^{1,}$\footnotemark[1] \\
$^{1}$ Department of Electronic Engineering, The Chinese University of Hong Kong\\
$^{2}$ SenseTime Group Limited\\
$^{1}${\tt\small \{ytshen, xiaotong, hsli,  xgwang\}@ee.cuhk.edu.hk  }\\
$^{2}${\tt\small yishuai@sensetime.com  } \\
}


\maketitle

\footnotetext[1]{Corresponding authors}
\begin{abstract}
	Vehicle re-identification is an important problem and has many applications in video surveillance and intelligent transportation. It gains increasing attention because of the recent advances of person re-identification techniques. However, unlike person re-identification, the visual differences between pairs of vehicle images are usually subtle and even challenging for humans to distinguish. Incorporating additional spatio-temporal information is vital for solving the challenging re-identification task. Existing vehicle re-identification methods ignored or used over-simplified models for the spatio-temporal relations between vehicle images. In this paper, we propose a two-stage framework that incorporates complex spatio-temporal information for effectively regularizing the re-identification results. Given a pair of vehicle images with their spatio-temporal information, a candidate visual-spatio-temporal path is first generated by a chain MRF model with a deeply learned potential function, where each visual-spatio-temporal state corresponds to an actual vehicle image with its spatio-temporal information. A Siamese-CNN+Path-LSTM model takes the candidate path as well as the pairwise queries to generate their similarity score. Extensive experiments and analysis show the effectiveness of our proposed method and individual components.
\end{abstract}

\section{Introduction}

Vehicle recognition is an active research field in computer vision, which includes applications such as vehicle classification~\cite{yang2015large,sochor2016boxcars,liu2016deepfine}, vehicle detection~\cite{matei2011vehicle}, and vehicle segmentation~\cite{mahendran2015car}. Vehicle re-identification (re-ID), which aims at determining whether two images are taken from the same vehicle, has recently drawn increasing attention from the research community ~\cite{feris2012large,zapletal2016vehicle,liu2016large}. It has important applications in video surveillance, public security, and intelligent transportation.

\begin{figure}[t]
\centering
\begin{tabular}{c@{\hspace{0mm}}c}
   &\includegraphics[scale=0.4]{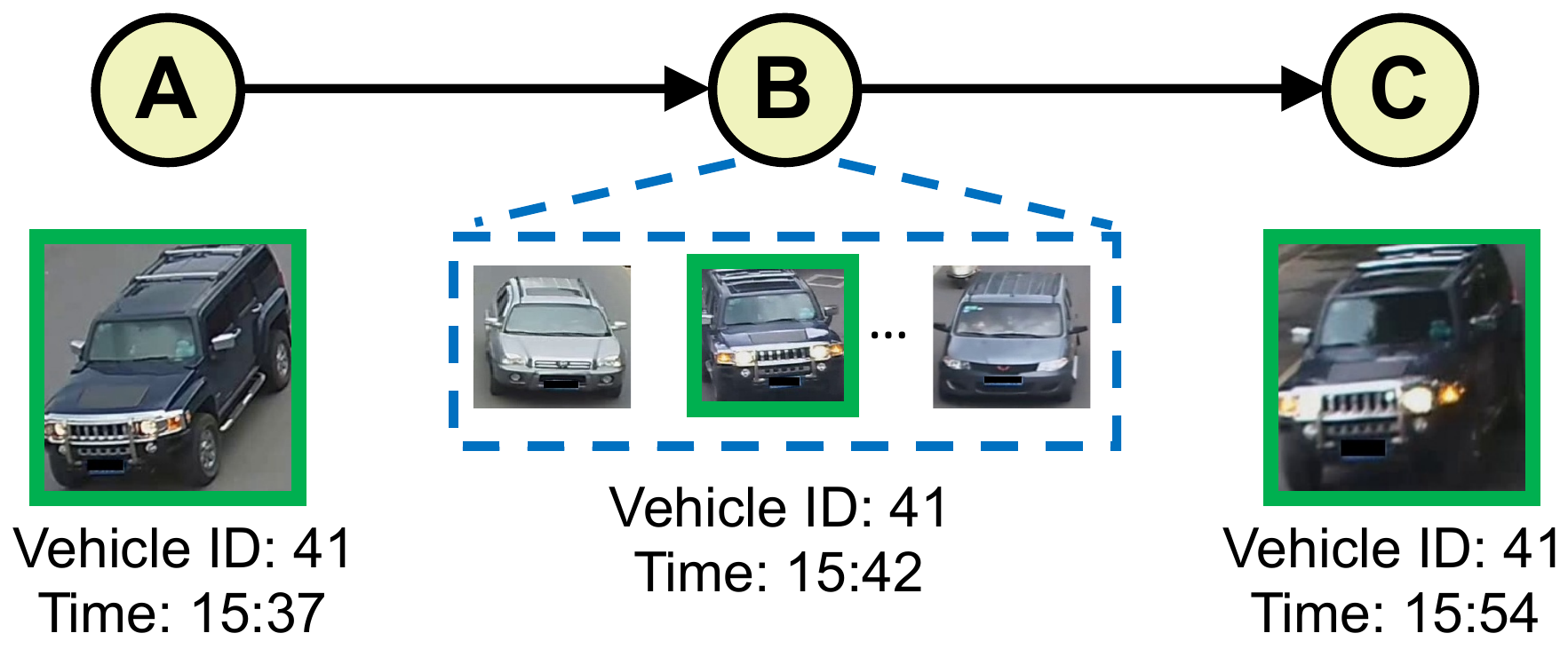}\\
   &(a) Similar vehicle observed at $B$ \\
   &\includegraphics[scale=0.4]{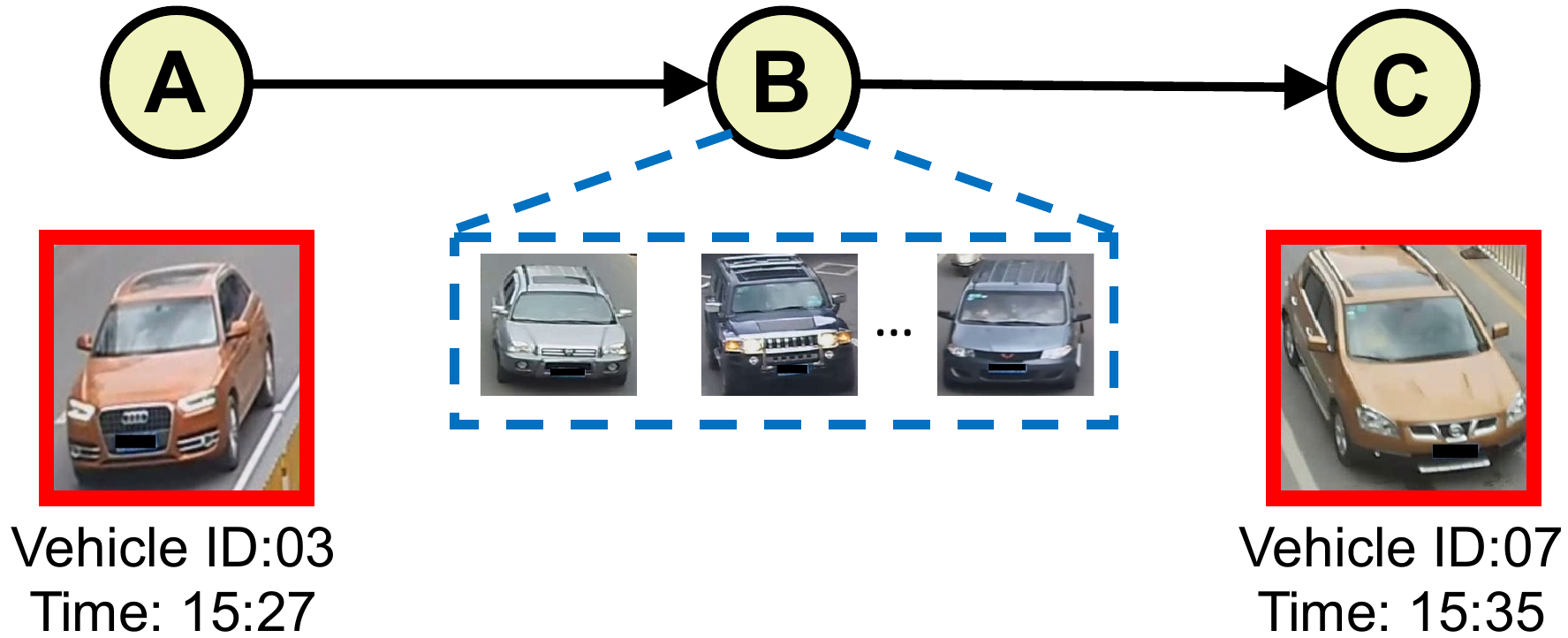}\\
   &(b) No similar vehicle observed at $B$
\end{tabular}
\vspace{-11pt}
   \caption{Illustration of spatio-temporal path information as important prior information for vehicle re-identification. (a) For vehicles with the same ID at $A$ and $C$, it has to be observed at $B$. (b) If a vehicle with similar appearance and proper time is not observed at $B$, vehicles at $A$ and $C$ are unlikely to be the same vehicle.}
\label{fig:path}
\end{figure}
Derived from person re-identification algorithms～\cite{ahmed2015improved, cheng2016person,hamdoun2008person}, most existing vehicle re-identification approaches~\cite{feris2012large,zapletal2016vehicle,liu2016large} rely only on appearance information. Such a problem setting is particularly challenging, since different cars could have very similar colors and shapes, especially for those belonging to the same manufacturer. Subtle cues for identification, such as license plates and special decorations, might be unavailable due to non-frontal camera viewpoints, low resolution, or poor illumination of the vehicle images. Therefore, it is not practical to use only appearance information for accurate vehicle re-identification.

To cope with such limitations, there are preliminary attempts on incorporating spatio-temporal information of the input images for more accurate vehicle re-identification. In \cite{liu2016deep}, Liu \etal utilized the time and geo-location information for each vehicle image. A spatio-temporal affinity is calculated between every pair of images. However, it favors pairs of images that are close to each other in both spatial and temporal domains. Such a spatio-temporal regularization is obviously over-simplified. More importantly, vital spatio-temporal path information of the vehicles provided by the dataset is ignored. The necessity of such spatio-temporal path prior is illustrated in Figure \ref{fig:path}. If a vehicle is observed at both camera $A$ and $C$, the same vehicle has to appear at camera $B$ as well. Therefore, given a pair of vehicle images at location $A$ and $C$, if an image with similar appearance is never observed at camera $B$ at a proper time, their matching confidence should be very low.

%

In this paper, we propose to utilize such spatio-temporal path information to solve the problem. The main contribution of our method is two-fold. (1) We propose a two-stage framework for vehicle re-identification. It first proposes a series of candidate visual-spatio-temporal paths with the query images as the starting and ending states. In the second stage, a Siamese-CNN+Path-LSTM network is utilized to determine whether each query pair has the same vehicle identity with the spatio-temporal regularization from the candidate path. In this way, all the visual-spatio-temporal states along the candidate path are effectively incorporated to estimate the validness confidence of the path. Such information is for the first time explored for vehicle re-identification. (2) To effectively generate visual-spatio-temporal path proposals, we model the paths by chain MRF, which could be optimized efficiently by the max-sum algorithm. A deep neural network is proposed to learn the pairwise visual-spatio-temporal potential function.

\section{Related Works}

\textbf{Vehicle re-identification.} Because of the quick advances of person re-identification approaches, vehicle re-identification started to gain attentions in recent years.
Feris \etal ~\cite{feris2012large} proposed an approach on attribute-based search of vehicles in surveillance scenes. The vehicles are classified by different attributes such as car types and colors. The retrieval is then conducted by searching vehicles with similar attributes in the database. Dominik \etal ~\cite{zapletal2016vehicle} utilized 3D bounding boxes for rectifying car images and then concatenated  color histogram features of pairs of vehicle images. A binary linear SVM is trained to verify whether the pair of images have the same identity or not. Liu \etal ~\cite{liu2016large,liu2016deep} proposed a vehicle re-identification dataset VeRi-776 with a large number of cars captured by 20 cameras in a road network. Vehicle appearances, spatio-temporal information and license plates are independently used to learn the similarity scores between pairs of images. For the appearance cues, a deep neural network is used to estimate the visual similarities between vehicle images.
Other vehicle recognition algorithms mainly focused on fine-grained car model classification~\cite{sochor2016boxcars,liu2016deepfine,yang2015large} instead of identifying the vehicles with the same or different identities.

\textbf{Deep neural networks.} \ In recent years, convolutional deep neuron networks have shown their effectiveness in large-scale image classification~\cite{krizhevsky2012imagenet}, object detection~\cite{DBLP:journals/corr/LiLOW17} and visual relationship detection~\cite{ livip}. For sequential data, the family of Recurrent Neural Networks including Long-Short Term Memory Network(LSTM)~\cite{hochreiter1997long} and Gated Recurrent Neural Network~\cite{cho2014properties} have achieved great success for tasks including image captioning~\cite{karpathy2015deep}, speech recognition~\cite{graves2014towards}, visual question answering~\cite{bahdanau2014neural}, person search~\cite{li2017person}, immediacy prediction~\cite{chu2015multi}, video classification~\cite{yue2015beyond} and video detection~\cite{kang2017tpn}. These works show that RNN is able to capture the temporal information in the sequential data and learn effective temporal feature representations, which inspires us to use LSTM network for learning feature representations for classifying visual-spatio-temporal paths.

\textbf{Person re-identification.} \ Person re-identification is a challenging problem that draws increasing attention in recent years \cite{ ahmed2015improved, cheng2016person, hamdoun2008person, wang2007shape}. 
State-of-the-art person re-identification methods adopted deep learning techniques.
Ahmed \etal \cite{ahmed2015improved} designed a pairwise verification CNN model for person re-identification with a pair of cropped pedestrian images as input and employed a binary verification loss function for training. Xiao \etal ~\cite{xiao2016learning,xiaoli2017joint} trained CNN with classification loss to learn the deep feature of person. Ding \etal ~\cite{ding2015deep} and Cheng \etal~ \cite{cheng2016person} trained CNN with triplet samples and minimized feature distances between the same person and maximize the distances between different people. Besides the feature learning, a large number of metric learning methods for person re-identification were also proposed~\cite{paisitkriangkrai2015learning, mcfee2010metric, koestinger2012large, weinberger2009distance, yi2014deep}. For person re-identification in multi-camera system, Hamdoun \etal ~\cite{hamdoun2008person} proposed an approach that matches signatures based on interest-points descriptors collected on short video sequences for person re-identification scheme in multi-camera surveillance systems.

\textbf{Spatio-temporal relations.} \ Spatio-temporal relations are widely exploited for objects association in multi-camera systems \cite{hamdoun2008person, wu2016model, cucchiara2007multi, kettnaker1999bayesian, javed2008modeling, ellis2003learning, neumann2002spatio},  Ellis \etal ~\cite{ellis2003learning} presented a method to learn both the
topological and temporal transitions from trajectory data which are obtained independently from single view target tracking in a multi-camera network.  Neumann \etal ~\cite{neumann2002spatio} presented an approach that combines the structure and motion estimation in a unified framework to recover an accurate 3D spatio-temporal description of an object. Loy \etal ~\cite{loy2009multi} proposed an approach for multi-camera activity correlation analysis which estimates the spatial and temporal topology of the camera network.

\begin{figure*}[t]
\begin{center}
   \includegraphics[width=0.7\linewidth]{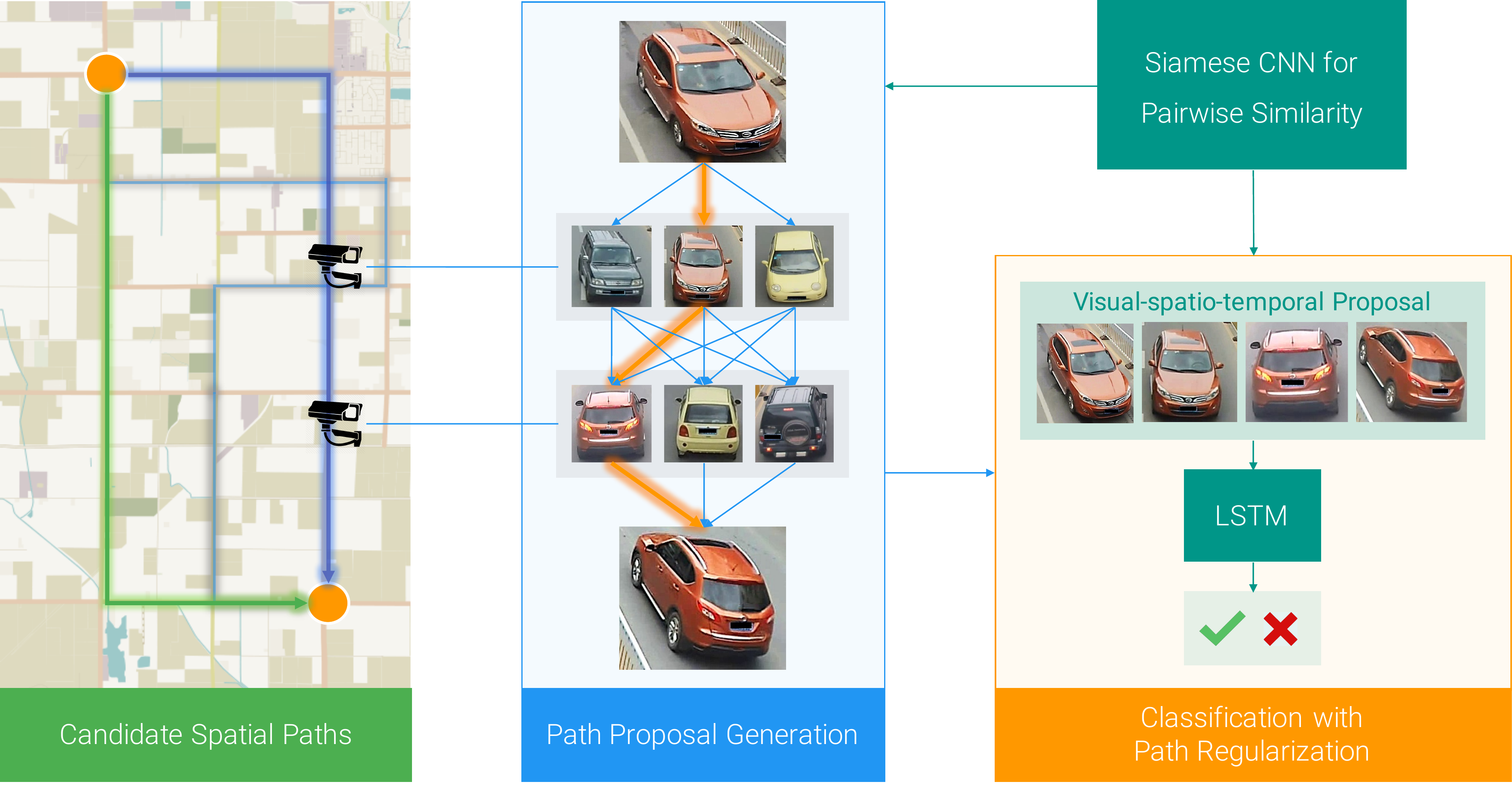}
\end{center}
\vspace{-18pt}
   \caption{Illustration of the overall framework. Given a pair of vehicle images, the visual-spatio-temporal path proposal is generated by optimizing a chain MRF model with a deeply learned potential function.
   The path proposal is further validated by the Path-LSTM and regularizes the similarity score by Siamese-CNN to achieve robust re-identification performance.}
\label{fig:hframe}
\end{figure*}

\section{Approach}

For vehicle re-identification with computer vision, cameras in a road network are needed to capture images of passing-by vehicles. Due to factors including inappropriate camera viewpoints, low resolution of the images and motion blurs of vehicles, car plate information might not always be available for solving the task.
 Given a pair of vehicle images with their spatio-temporal information, the similarity score between the two vehicle images is needed to determine whether the two images have the same identity. 
Each image is associated with three types of information, \ie, visual appearance, the timestamp, and the geo-locatoin of the camera. We call such information the \emph{visual-spatio-temporal state} of the vehicle images. Our proposed approach takes two visual-spatio-temporal states as input and outputs their similarity score with a two-stage framework, which is illustrated in Figure \ref{fig:hframe}. In stage 1, instead of just considering the simple pairwise relations between two queries, our proposed approach first generates a candidate visual-spatio-temporal path with the two queries as starting and ending states. In stage 2, the candidate visual-spatio-temporal path acts as regularization priors and a Siamese-CNN+path-LSTM network is utilized to determine whether the queries have the same identity.



\subsection{Visual-spatio-temporal path proposals}

Given a pair of queries, existing vehicle or person re-identification algorithms mainly considers the pairwise relations between the queries, \eg, the compatibility of the visual appearances and spatio-temporal states of the two queries. As illustrated in Figure \ref{fig:path}, such pairwise relations are usually over-simplified and cannot effectively distinguish difficult cases in practice. The visual-spatio-temporal path information could provide vital information for achieving more robust re-identification. 
The problem of identifying candidate visual-spatio-temporal paths is modeled as chain Markov Random Fields (MRF). 
Given a pair of queries, candidate spatial paths are first identified with the their geo-locations as starting and ending locations. 
The visual-spatio-temporal states along the spatial paths are then optimized with the deeply learned pairwise potential function to generate candidate visual-spatio-temporal paths.


To obtain candidate spatial paths for a pair of starting and ending locations, all possible spatial paths that the same vehicle has passed by  are collected from the training set. For a large road network, the multiple candidate spatial paths between every pair of locations could be pre-collected.


\subsubsection{Chain MRF model for visual-spatio-temporal path proposal}
\label{cmrf}
From the set of candidate spatial paths, our approach proposes one candidate visual-spatio-temporal path for regularizing the vehicle re-identification result. The problem of identifying visual and temporal states along the candidate spatial paths is modeled as optimizing chain MRF models.


Let $N$ denote the number of cameras on a candidate spatial path, where each of the $N$ cameras is associated with one of the  random variables $\mathbf{X} = \{X_1, X_2, \cdots, X_N\}$ on a chain MRF model. For the $i$-th random variable (camera) $X_i$, its domain is the set of all $k$ visual-spatio-temporal states (all $k$ images with their spatio-temporal information) at this camera, $S_i  = \{s_{i,1}, \cdots, s_{i, k} \}$, where $s_{i,j}$= $\{I_{i,j}, t_{i,j}, l_i\}$ is a triplet of the $j$th visual image at the $i$th camera, its timestamp $t_{i,j}$, and the camera location $l_i$.

Let $p$ and $q$ represent the visual-spatio-temporal states of the two queries. Obtaining the optimal visual-spatio-temporal path based on a candidate spatial path can be achieved by maximizing the following distribution,
\begin{align}
p(\mathbf{x}|x_1 & = p, x_N = q) = \nonumber \\
&\frac{1}{Z} \psi(p, x_2)  \psi(x_{N-1}, q) \prod_{i=2}^{N-2}{\psi(x_i, x_{i+1}}),
\label{eq:mrf}
\end{align}
where $\psi (x_i, x_{i+1})$ is the pairwise potential function of $x_i$ and $x_{i+1}$ being of the same car. Ideally, if $x_i$ and $x_{i+1}$ denote the states with the same vehicle identity, a proper potential function would have a large value, while small otherwise. The $\psi$ function is learned as a deep neural network and is introduced in details in the next subsection.
 

The above probability needs to be maximized with proper time constraints,
\begin{align}
&\mathbf{x^*} = \mathop{\arg\max}\limits_\mathbf{x} ~p(\mathbf{x} | x_1 = p, x_N = q),\\
\textnormal{sub}&\textnormal{ject to} ~~~~t_{i,{k}_i^*} \leq t_{i+1,{k}_{i+1}^*} ~~\forall i \in \{1, \cdots, N-1\},
\end{align}
where ${k}_i^*$ and ${k}_{i+1}^*$ represent the indices of the optimal visual-spatio-temporal states for $x_i$ and $x_{i+1}$, respectively. The above constraints state that the obtained visual-spatio-temporal path must be feasible in time, \ie, the timestamps of vehicle images must be keep increasing along the path.


 
The distribution can be efficiently optimized by the max-sum algorithm, which is equivalent to dynamic programming for chain models \cite{cormen2009introduction}. 
The maximum of the probability can be written as,
\begin{align}
&~~~~~\mathop{\max}\limits_\mathbf{x} ~p(\mathbf{x}|x_1 = p, x_N = q)\\
&= \frac{1}{Z}\psi(p, x_2)\psi(x_{N-1}, q)\mathop{\max}\limits_{{x_{2}}}\cdots \mathop{\max}\limits_{x_{N-1}} \prod_{i=2}^{N-1}\psi ({x_i, x_{i+1})}\\
&=\frac{1}{Z}\mathop{\max}\limits_{x_{2}} \left[\psi(p,x_2)\psi (x_2, x_3) \left[\cdots \mathop{\max}\limits_{x_{N-1}}\psi (x_{N-1}, x_{q}) \right] \cdots \right]
\end{align}

After obtaining the optimal state for each random variable (camera) on the candidate spatial path, the candidate visual-spatio-temporal path is generated. 

For a pair of queries, multiple candidate visual-spatial-temporal paths are obtained. we define an empirical averaged potential for the optimed solution of each candidate path,
 \begin{align}
S(\mathbf{x^*}) = \frac{1}{N-1} \left( \psi(p, 2) + \sum_{i = 2}^{N-2}\psi(x_i^*, x_{i+1}^*) + \psi(x_{N-1}^*, q) \right),
\label{eq:empirical_average}
\end{align}
where $1/(N-1)$ normalizes the overall confidence for candidate visual-spatio-temporal paths with different lengths. Then we choose the visual-spatio-temporal proposal among the candidate paths.

\begin{figure}[t]
\begin{center}
   \includegraphics[scale=0.4]{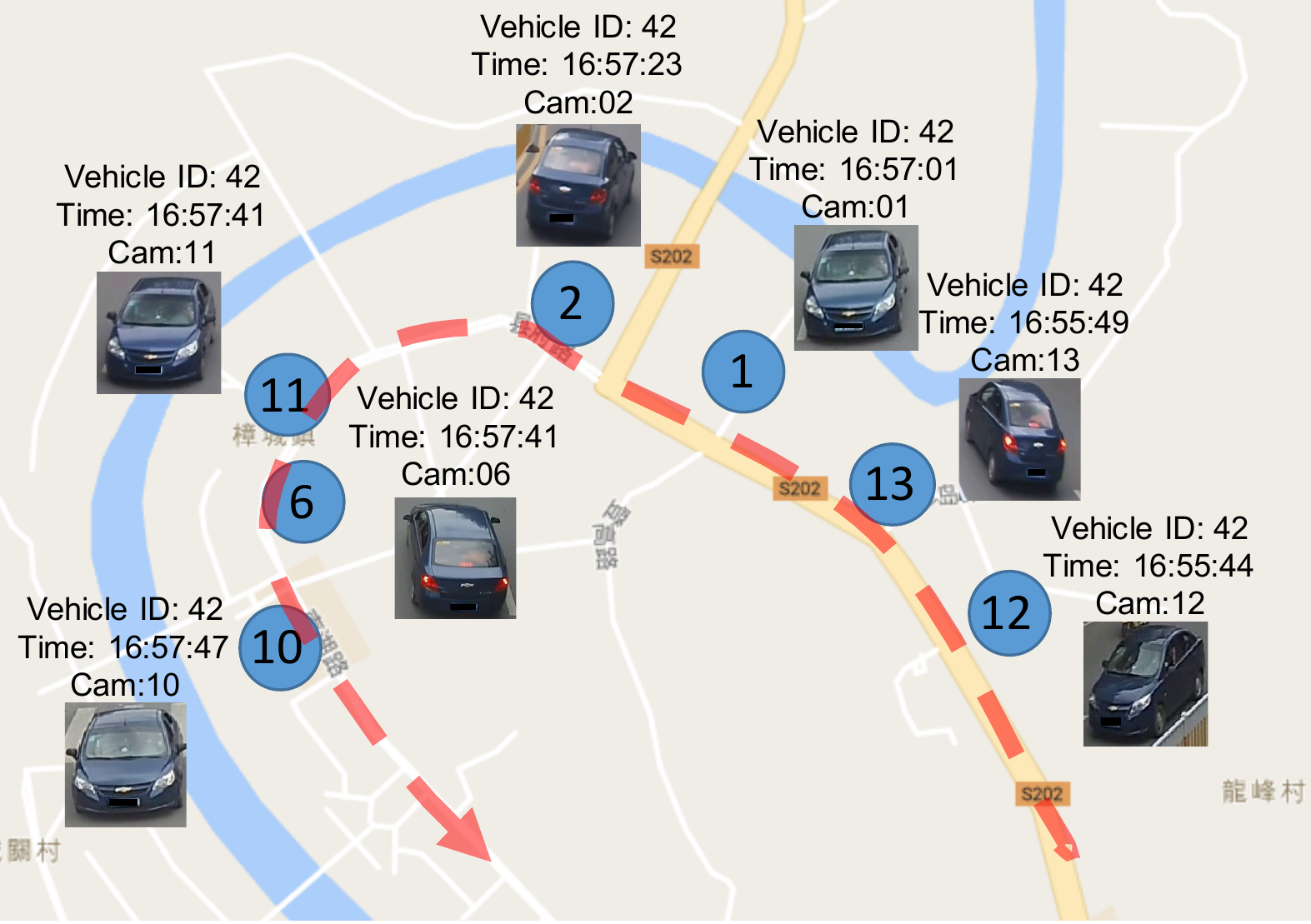}
\end{center}
\vspace{-16pt}
   \caption{An example visual-spatio-temporal path proposal on the VeRi dataset \cite{liu2016deep} by our chain MRF model.}
\label{fig:path_example}
\end{figure}

For every pair of queries, even if they do not have the same identity, the proposed algorithm always tries to generate the most feasible path in terms of both visual and spatio-temporal compatibility between neighboring cameras as the path proposals. Some examples of the candidate visual-spatio-temporal paths are shown in Figure \ref{fig:path_example}.

For efficiency, when generating visual-spatio-temporal paths for all state pairs in the dataset, our method utilizes a systematic way to avoid redundant computation. For in- stance, suppose cameras $A$ and $C$ have two possible paths $A-B_1-C$ and $A-B_2-C$. When calculating path proposals between queries on cameras $A$ and $C$, the sub-paths $A-B_1$ and $A-B_2$ are also computed during the computation process, and can be reused by other queries on $A-B_1$ and $A-B_2$. The details of time complexity analysis will be introduced in Section \ref{ssec:timeca}.

%


\begin{figure}[t]
\begin{center}
   \includegraphics[width=0.9\linewidth]{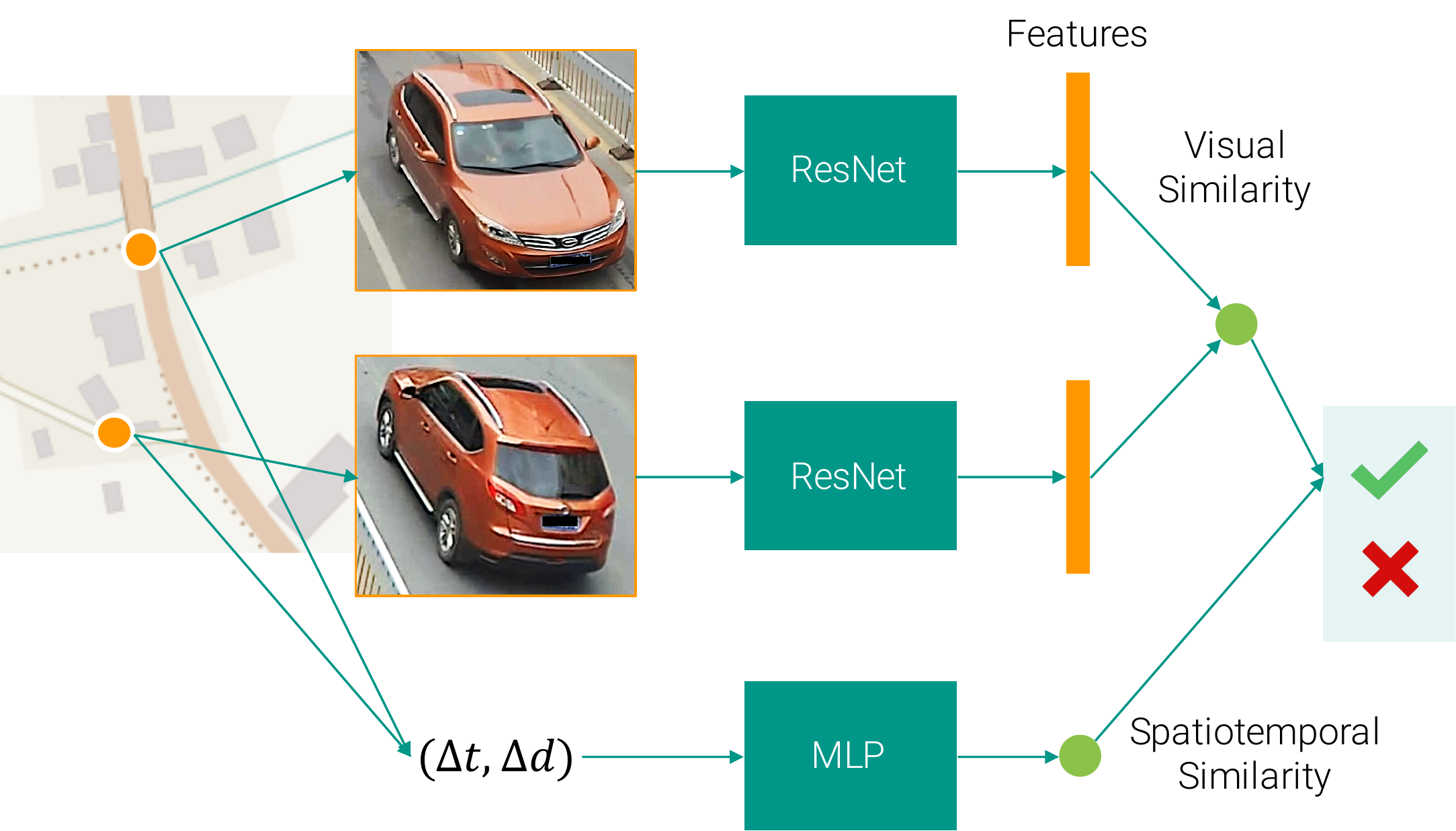}
\end{center}
	\vspace{-16pt}
   \caption{A Siamese-CNN is learned as the pairwise potential function for the chain MRF model, which takes a pair of visual-spatio-temporal states as inputs and estimates their pairwise similarity.}
\label{fig:siamese}
\end{figure}

\subsubsection{Deep neural networks as pairwise potential functions}
\label{ssec:potential}

The pairwise potential function $\psi (x_i, x_{i+1})$ in (\ref{eq:mrf}) evaluates the compatibility between two visual-spatio-temporal states for neighboring random variables $x_i$ and $x_{i+1}$. We learn the potential function $\psi$ as a two-branch deep neural network, whose structure is illustrated in Figure \ref{fig:siamese}. The visual branch and spatio-temporal branch estimate pairwise compatibility between pairwise visual and spatio-temporal states.

The visual branch (Siamese-Visual) is designed as a Siamese network with a shared ResNet-50 \cite{he2016deep}. It takes two images $I_{i,k}$ and $I_{i+1,j}$ at cameras $x_i$ and $x_{i+1}$ as inputs and utilizes features from the ``global pooling'' layers to describe their visual appearances. The visual similarity between the two images is computed as the inner-product of the two ``global pooling'' features followed by a sigmoid function.

The other branch computes the spatio-temporal compatibility. Given the timestamps $\{t_{i,k}, t_{i+1,j}\}$ and the two geo-locations $\{l_i, l_{i+1}\}$ of at cameras $i$ and $i+1$, the input features of the branch are calculated as their time difference and spatial difference,
\begin{align}
&\Delta t_{i,i+1}(k,j) = t_{i+1,j} - t_{i,k},\\
&\Delta d_{i,i+1} = |l_{i+1} - l_{i}|,
\end{align}
where $t_{i,k}$ denotes the timestamp of the $k$-th state at camera $i$.
The scalar spatio-temporal compatibility is obtained by feeding the concatenated features, $[\Delta t_{i,i+1}(k,j), \Delta d_{i,i+1}]^T$, into a Multi-Layer Perception (MLP) with two fully-connected layers and a ReLU non-linearity function after the first layer and a sigmoid function after the second layer.

The outputs of the two branches are concatenated and input into a $2\times 1$ fully-connected layer with a sigmoid function to obtain the final compatibility between the two states, which takes all visual, spatial and temporal information into consideration.

For training the pairwise potential network, Siamese-CNN, we first pretrain the ResNet-50 network to classify vehicle identity with the classification cross-entropy loss function.
All pairs of visual-spatio-temporal states at neighboring random variables (cameras) are then collected for finetuning the whole network. If a pair has the same vehicle identities, they are treated as positive samples, while the pairs with different identities are treated as negative ones. The positive-to-negative sampling ratio is set to 1:3. The two-branch network is trained with a 0-1 cross-entropy loss function and stochastic gradient descent.


\begin{figure}[t]
\centering
\begin{tabular}{c@{\hspace{-0mm}}c}
   &\includegraphics[scale = 0.25]{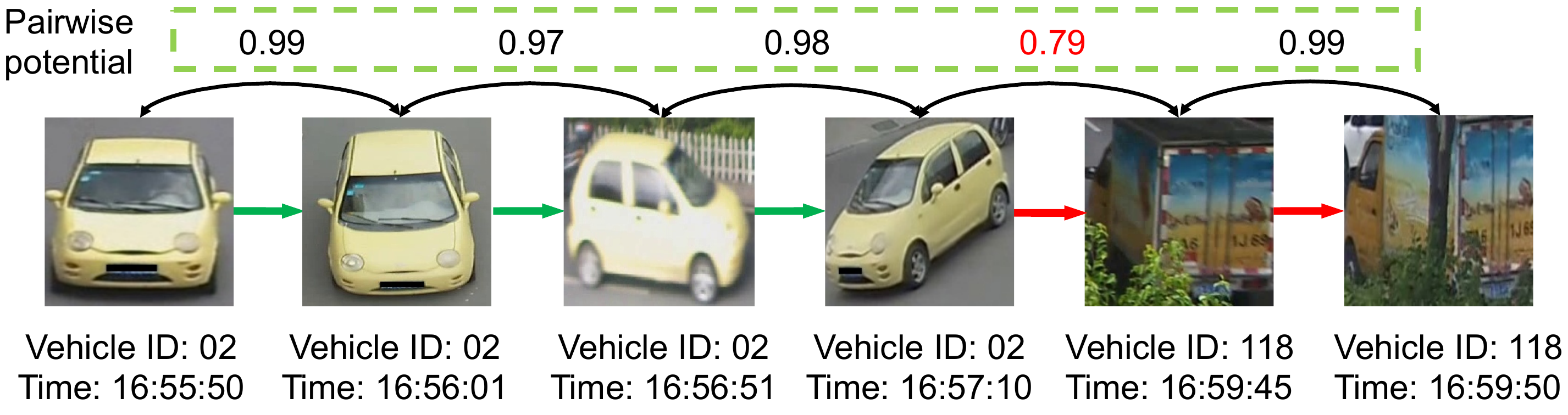}\\
   & (a) Invalid path. Empirical averaged potential: 0.946\\
   &\includegraphics[scale = 0.25]{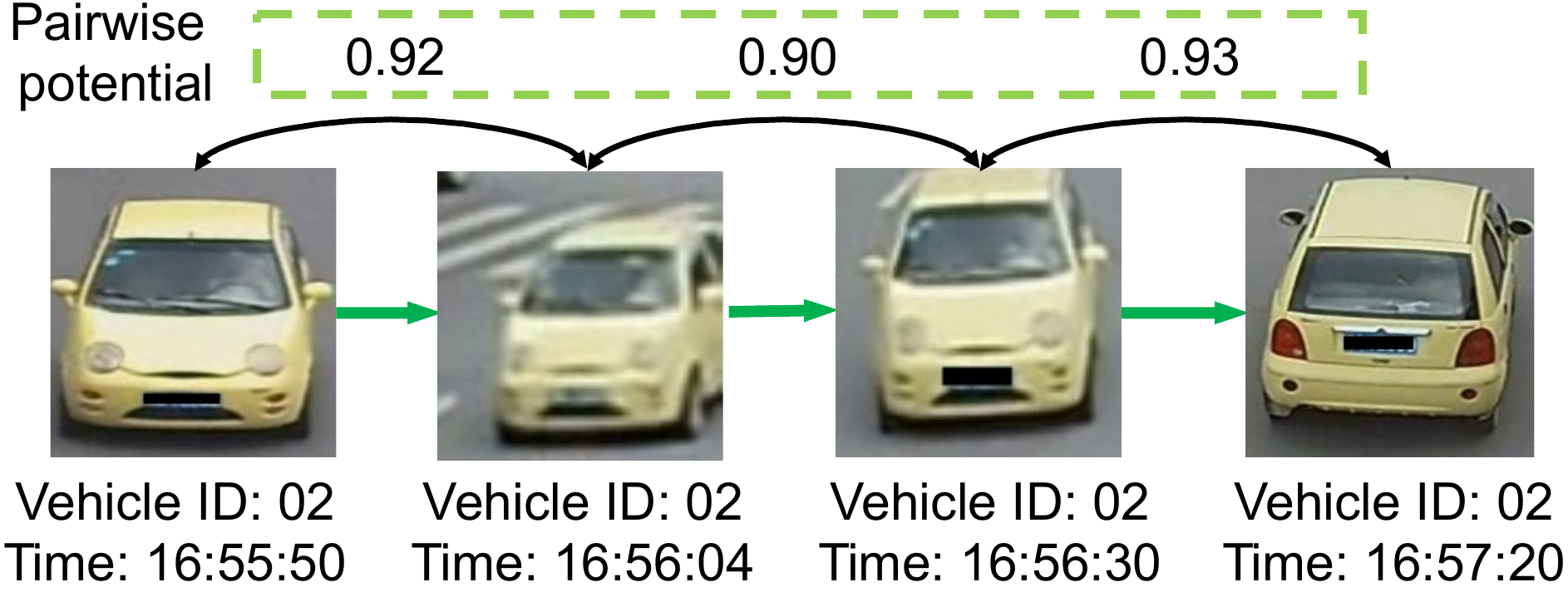}\\
   & (b) Valid path. Empirical averaged potential: 0.916
\end{tabular}
\vspace{-12pt}
   \caption{Examples of empirical averaged potential favoring longer paths. The invalid longer path in (a) has a higher averaged potential than the valid path in (b).}
\label{fig:flush}
\end{figure}

\subsection{Siamese-CNN+Path-LSTM for query pair classification}
\label{ssec:lstm}

Our proposed candidate visual-spatio-temporal path proposal algorithm generates the most feasible path for each query pair, even if they do not have the same identity. One naive solution of ranking the similarities of query pairs would be directly treating their maximum probability  (Eq. (\ref{eq:mrf})) or the empirical averaged potential (Eq. (\ref{eq:empirical_average})) as the final similarity scores for ranking. However, there are limitations when either of the strategies is adopted. For calculating the maximum probability in Eq. (\ref{eq:mrf}), the partition function $Z$ needs to be calculated, which is generally time-consuming. For the empirical averaged potential in Eq. (\ref{eq:empirical_average}), it is biased and would favor longer paths. An exmaple is illustrated in Figure \ref{fig:flush}. Given two pairs of negative queries with different path lengths, since the path proposal algorithm tries to generate most feasible paths for both pairs, there might be only one identity switch along each path. The empirical averaged confidence for the longer path would be higher because the low pairwise confidence would be diminished by a larger $N$.

\begin{figure}[t]
\begin{center}
   \includegraphics[width=0.8\linewidth]{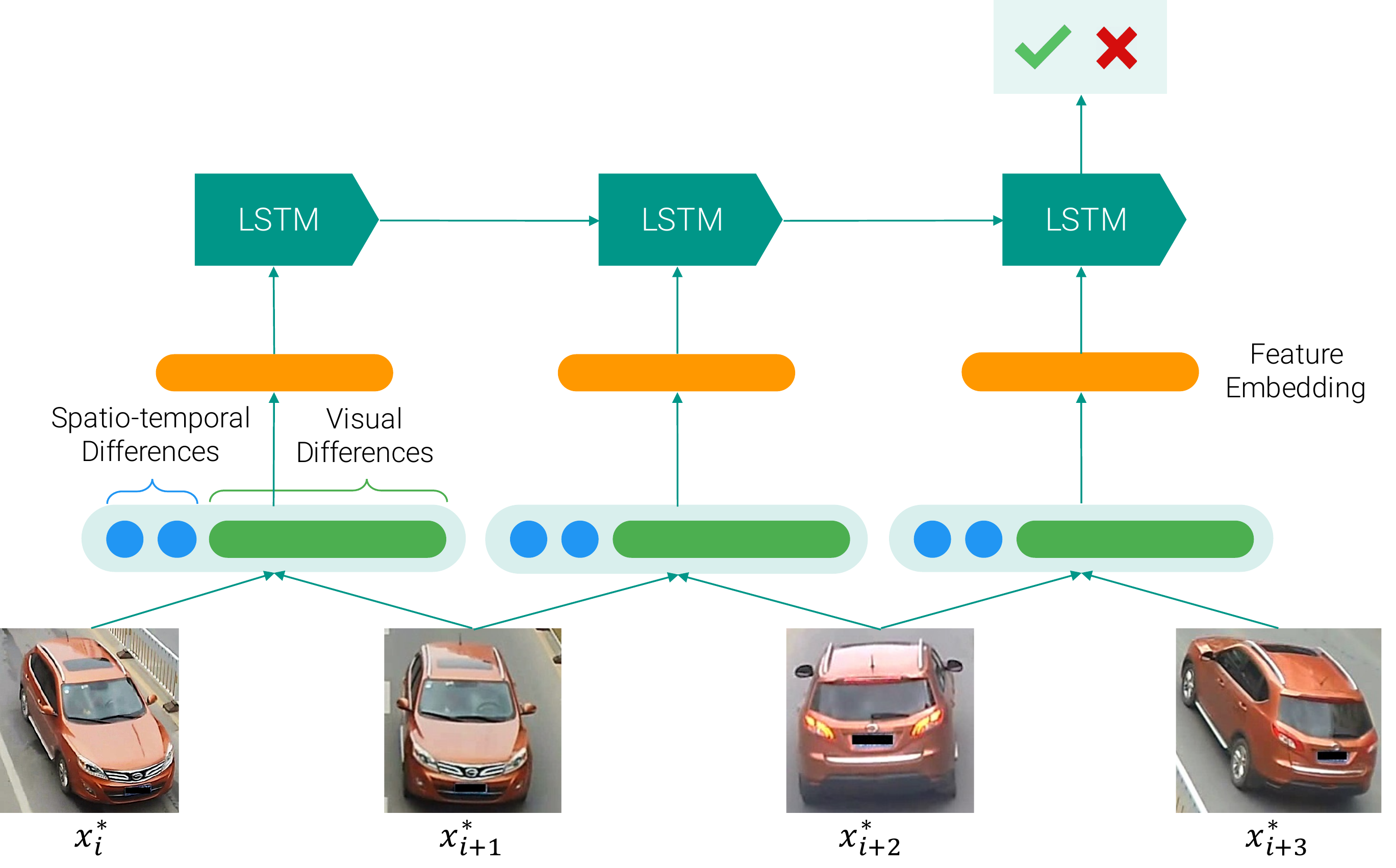}
\end{center}
\vspace{-18pt}
   \caption{The network structure of the Path-LSTM. It takes visual and spatio-temporal differences of neighboring states along the path proposal as inputs, and estimates the path validness score.}
\label{fig:LSTM}
\end{figure}

Given a pair of queries, we utilize their candidate visual-spatio-temporal path as priors to determine whether the query pair has the same identity or not with a Siamese-CNN+path-LSTM network. The network structure is illustrated in Figure \ref{fig:LSTM}. where the Siamese-CNN has the same structure as the overall network in Section \ref{ssec:potential}. It directly takes the query pair as input and estimates the similarity between the queries.

For the candidate visual-spatio-temporal path, a path-LSTM is adopted to judge whether the path is valid or not. The framework of path-LSTM network is shown in Figure \ref{fig:LSTM}. Each step of the path-LSTM processes a step along the candidate path and the length of the LSTM is therefore $N-1$. At each step, the input features of the path-LSTM, $y_i$, is the concatenation of visual difference ${\Delta I^*}_{i,i+1}$, spatial difference $\Delta d_{i,i+1}^*$, and temporal difference ${\Delta t}_{i,i+1}^*$ between the visual-spatio-temporal states at cameras $i$ and $i+1$, followed by a fully-connected layer with 32 output neurons and a ReLU non-linearity function, \ie, $y_i = f( [ {\Delta I}_{i,i+1}^*, \Delta d_{i,i+1}^*, {\Delta t}_{i,i+1}^* ]^T)$, where $f$ denotes the feature transformation by the fully-connected layer.

Given the visual-spatio-temporal states on the candidate path, the image $I_{i, k_i^*}$ and its associated time stamp $t_{i, {k}_i^*}$ is fixed at camera $i$, where ${k}_i^*$ denotes the optimized index of the visual-spatio-temporal state. The visual difference ${\Delta I}^*_{i,i+1}$ is calculated as absolute difference between the visual features of $I_{i, {k}_i^*}$ and $I_{i+1, {k}_{i+1}^*}$, which are obtained as the ResNet-50 global pooling features followed by a 32-neuron fully-connected layer and a ReLU function,
\begin{align}
	{\Delta I}_{i,i+1}^* = \left| R \left(I_{i, {k}_i^*}\right) - R\left(I_{i+1, {k}_{i+1}^*} \right) \right|,
\end{align}
where $R$ denotes the feature transformation for the input images. The spatial difference $\Delta d_{i,i+1}$ is calculated as $\Delta d_{i,i+1} = l_{i+1} - l_i$, and the temporal difference is
\begin{align}
	{\Delta t}_{i,i+1} =  t_{i+1, {k}_{i+1}^*} - t_{i, {k}_i^*}	.
\end{align}


The LSTM consists of a memory cell $c_t$ and three control gates: input gate $ig$, output gate $og$ and forget gate $fg$ at each time step $t$. With the input feature $y_t$, The LSTM updates the memory cell $c_t$ and hidden state $h_t$ with the following equations,
\begin{align}
fg_t & = \sigma(W_f \cdot [h_{t-1}, y_t]) + b_f )\\
ig_t & = \sigma(W_i \cdot [h_{t-1}, y_t]) + b_i )\\
og_t & = \sigma(W_o \cdot [h_{t-1}, y_t]) + b_o )\\
\tilde{c}_t &= \tanh (W_c \cdot [h_{t-1}, y_t] + b_c) \\
c_t &= fg_t \ast c_{t-1} + ig_t \ast \tilde{c}_t \\
h_t &= og_t \ast \tanh (c_t)
\end{align}
where $\ast$ represents the element-wise multiplication, $W$ and $b$ are the parameters.

The number of hidden neurons of our Path-LSTM is set to 32. The hidden feature of the Path-LSTM at the last step is fed into a fully-connected layer to obtain the valid-path confidence score, which is added with the pairwise similarity score generated by the Siamese-CNN to represent the final similarity score. In this way, the Path-LSTM provides important regularization for estimating final matching similarity.


Both the Siamese-CNN and Path-LSTM are pretrained separately and then finetuned jointly. Their training samples are prepared similarly to those for learning the pairwise potential function in Section \ref{ssec:potential}. However, the training samples here are no longer restricted to only visual-spatio-temporal states from neighboring cameras but from any camera pair in the entire camera network. The Path-LSTM is first pretrained with the Adam algorithm \cite{kingma2014adam}. The whole Siamese-CNN+Path-LSTM network is then finetuned in an end-to-end manner with stochastic gradient descent and 0-1 cross-entropy loss function.



\section{Experiments}


\subsection{Dataset and evaluation metric}
\label{ssec:dataset}
For evaluating the effectiveness of our vehicle re-identification framework, we conduct experiments on the VeRi-776 dataset~\cite{liu2016deep}, which is the only existing vehicle re-identification dataset  providing spatial and temporal annotations. The VeRi-776 dataset contains over 50,000 images of 776 vehicles with identity annotations, image timestamps, camera geo-locations, license plates, car types and colors information. Each vehicle is captured by 2 to 18 cameras in an urban area of 1$km^2$ during a 24-hour time period. The dataset is split into a train set consisting of 37,781 images of 576 vehicles, and a test set of 11,579 images belonging to 200 vehicles. A subset of 1,678 query images in the test set are used as to retrieve corresponding images from all other test images.

The mean average precision (mAP), top-1 accuracy and top-10 accuracy are chosen as the evaluation metric. Given each query image in the test image subset for retrieving other test images, the average precision for each query $q$ is calculated by
\begin{align}
	AP(q) = \frac{ \sum_{k=1}^{n} P(k) \times rel(k)}{N_{gt}}
\end{align}
where $P(k)$ denotes the precision at cut-off $k$, $rel(k)$ is an indication function equaling 1 if the item at rank $k$ is a matched vehicle image, zero otherwise, $n$ is the number for retrieval, and $N_{gt}$ denotes the number of ground truth retrievals for the query.
The mean average precision for all query images is then calculated by
\begin{align}
mAP = \frac{\sum_{q=1}^Q AP(q)}{Q},
\end{align}
where $Q$ is the number of all queries (1,678 for the dataset). Following the experimental setup in  Liu \etal \cite{liu2016deep}, for each query image, only images of the same vehicles from other cameras would be taken into account for calculating the mAP, top-1 and top-5 accuracies.

Since our proposed method generates one visual-spatio-temporal candidate path for each pair of query images, we can extend the evaluation metrics to the whole sequence of images. For every pair of query images that have the same vehicle identity, we obtain the vehicle's actual visual-spatio-temporal path and compare it with our candidate path using Jaccard Similarity \cite{jaccard1901etude},
\begin{align}
JS(P_p, P_g) = \frac{P_p \cap P_g}{P_p \cup P_g},
\end{align}
where $P_p$ is the set of retrieved images  on the proposal path between the query pair, and $P_g$ is the set of the groundtruth images between them. We further define the average Jaccard Similarity for all query images,
\begin{align}
AJS(P_p, P_g) = \frac{1}{Q} \sum_{q=1}^{Q}\frac{P_{p_q} \cap P_{g_q}}{P_{p_q} \cup P_{g_q}}.
\end{align}

\subsection{Compared With Vehicle Re-ID methods}
\label{ssec:comparison}

We compare our proposed approach with the state-of-the-art methods \cite{liu2016large,liu2016deep} and design several baselines on the VeRi-776 dataset to evaluate the effectiveness of our proposed method.

\begin{itemize}
\vspace{-7pt}
\item \emph{Siamese-CNN+Path-LSTM} denotes our final approach which takes pairs of query visual-spatio-temporal states for Siamese-CNN and proposed visual-spatio-temporal path for the Path-LSTM to generate the final similarity score. Note that our approach did not utilize the plate information in the images as that in \cite{liu2016deep}.
\vspace{-3pt}
\item \emph{FACT} and \emph{FACT+Plate-SNN+STR.} Liu \etal \cite{liu2016large} proposed the FACT model that combines deeply learned visual feature from GoogleNet \cite{szegedy2015going}, BOW-CN and BOW-SIFT feature to measure only the visual similarity between pairs of query images.
In \cite{liu2016deep}, they further integrated visual appearance, plate and spatio-temporal information for vehicle re-identification. For appearance similarity, the same FACT model is adopted. They utilized a Siamese Neural Network (Plate-SNN) to compare visual similarity between plate regions. The spatio-temporal similarity is computed as
\begin{align}
STR(i,j) = \frac{|T_i - T_j|}{T_{max}} \times \frac{\delta{(C_i - C_j)}}{D_{max}},
\label{eq:str}
\end{align}
where $T_i$ and $T_j$ are the timestamps of two queries, $\delta(C_i, C_j)$ is the space distance between two cameras, and $T_{max}$ and $D_{max}$ are the maximum time distance and space distance in the whole dataset.
\vspace{-3pt}
\item \emph{Siamese-Visual.} This baseline generates the similarity between a query pair with only pairwise visual information by using only the visual branch of the Siamese-CNN in Section \ref{ssec:lstm}. No spatio-temporal information is used for obtaining the similarity score.
\vspace{-3pt}
\item \emph{Siamese-Visual+STR.} Instead of learning spatio-temporal relations by deep neural networks, this baseline sums up the scores by the above \emph{Siamese-Visual} and the spatial-temporal relation score (STR, Eq. (\ref{eq:str}) proposoed in \cite{liu2016deep}. The weight between the two terms are manually searched for the best performance.
\vspace{-3pt}
\item \emph{Siamese-CNN.} The baseline is the same as the Siamese-CNN in Section \ref{ssec:lstm}. Compared with the above baseline, it uses both visual and spatio-temporal information of the two queries for determine their similarity score. Candidate visual-spatio-temporal paths are not used in this baseline.
\vspace{-3pt}
\item \emph{Chain MRF model.} After obtaining the candidate visual-spatio-temporal path for a query pair by the chain MRF model in Section \ref{cmrf}, we directly utilize the empirical average by Eq. (\ref{eq:empirical_average}) as the pairwise similarity of the query pair.
\vspace{-3pt}
\item \emph{Path-LSTM only.} The proposed Path-LSTM estimate the validness score of the proposed visual-spatio-temporal path. We test only use Path-LSTM result without combining it with the Siamese-CNN.
\vspace{-3pt}
\item \emph{Siamese-CNN-VGG16.} This is the same as Siamese-CNN but only replaces ResNet50 with VGG16.
\vspace{-3pt}
\item \emph{Path-LSTM-VGG16.} This is the same as Path-LSTM but only replaces Siamese-CNN with Siamese-CNN-VGG16. 
\vspace{-3pt}
\item \emph{Siamese-CNN-VGG16+Path-LSTM-VGG16.} This is as same as  Siamese-CNN+Path-LSTM but only replaces ResNet50 with VGG16.
\vspace{-3pt}
\end{itemize}

\begin{table}
\begin{center}
\begin{tabular}{|l|c|}
\hline
Method & mAP (\%) \\
\hline\hline
FACT \cite{liu2016large} & 18.49\\
FACT+Plate-SNN+STR \cite{liu2016deep} & 27.77 \\
Siamese-Visual & 29.48 \\
Siamese-Visual+STR & 40.26\\
Siamese-CNN &  54.21\\
Chain MRF model & 44.31\\
Path-LSTM & 54.49\\
Siamese-CNN-VGG16 & 44.32\\
Path-LSTM-VGG16 & 45.56\\
\tabincell{c}{Siamese-VGG16+\\PathLSTM-VGG16} & 46.85\\
Siamese-CNN+Path-LSTM & \textbf{58.27}\\
\hline
\end{tabular}
\end{center}
\vspace{-18pt}
\caption{mAP by compared methods on the VeRi-776 dataset \cite{liu2016deep}.}
\label{tab:mAP}
\end{table}

\begin{table}
\begin{center}
\begin{tabular}{|l|c|c|}
\hline
Method & top-1 (\%) & top-5 (\%) \\
\hline\hline
FACT \cite{liu2016large} &50.95 &73.48 \\
FACT+Plate-SNN+STR \cite{liu2016deep} &61.44  &78.78 \\
Siamese-Visual & 41.12 & 60.31\\
Siamese-Visual+STR & 54.23 & 74.97\\
Siamese-CNN & 79.32& 88.92\\
Chain MRF model & 54.41 & 61.50\\
Path-LSTM & 82.89 & 89.81\\
Siamese-CNN-VGG16 & 54.41 & 61.50\\
Path-LSTM-VGG16 & 47.79 & 62.63\\
\tabincell{c}{Siamese-VGG16+\\PathLSTM-VGG16} & 50.95 & 61.62\\
Siamese-CNN+Path-LSTM & \textbf{83.49} & \textbf{90.04}\\
\hline
\end{tabular}
\end{center}
\vspace{-18pt}
\caption{Top-1 and top-5 accuracies by compared methods on the VeRi-776 dataset \cite{liu2016deep}.}
\label{tab:top}
\end{table}

\begin{figure}[t]
\centering
\begin{tabular}{c@{\hspace{0mm}}c}
   &\includegraphics[scale = 0.33]{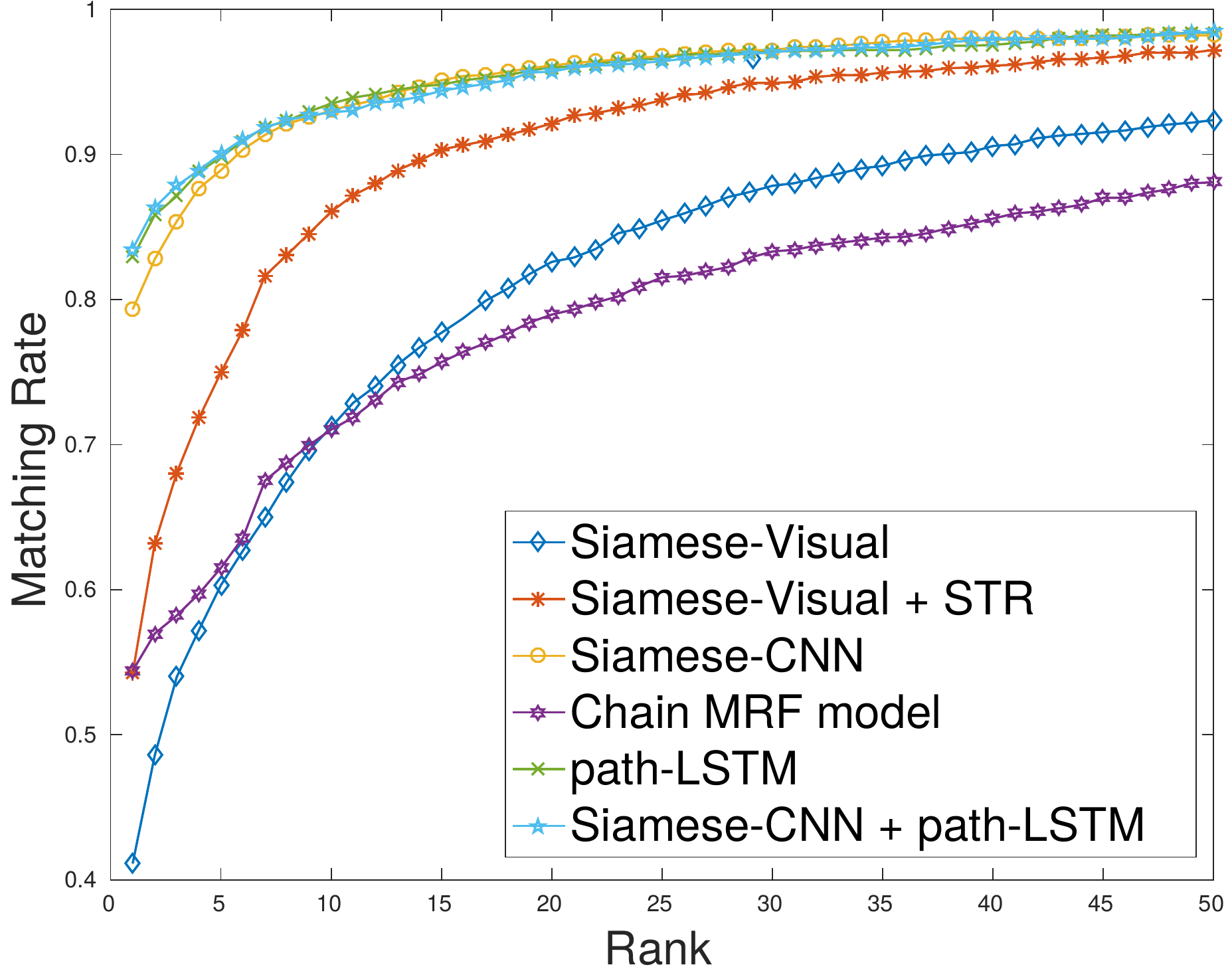}
\end{tabular}
	\vspace{-16pt}
   \caption{The CMC curves of different methods.}
\label{fig:CMC}
\end{figure}
 
\subsection{Experiment Results} 
\label{ssec:results}
\begin{figure}[t]
\begin{tabular}{c@{\hspace{2.5mm}}c}
\centering
   &\includegraphics[scale=0.27]{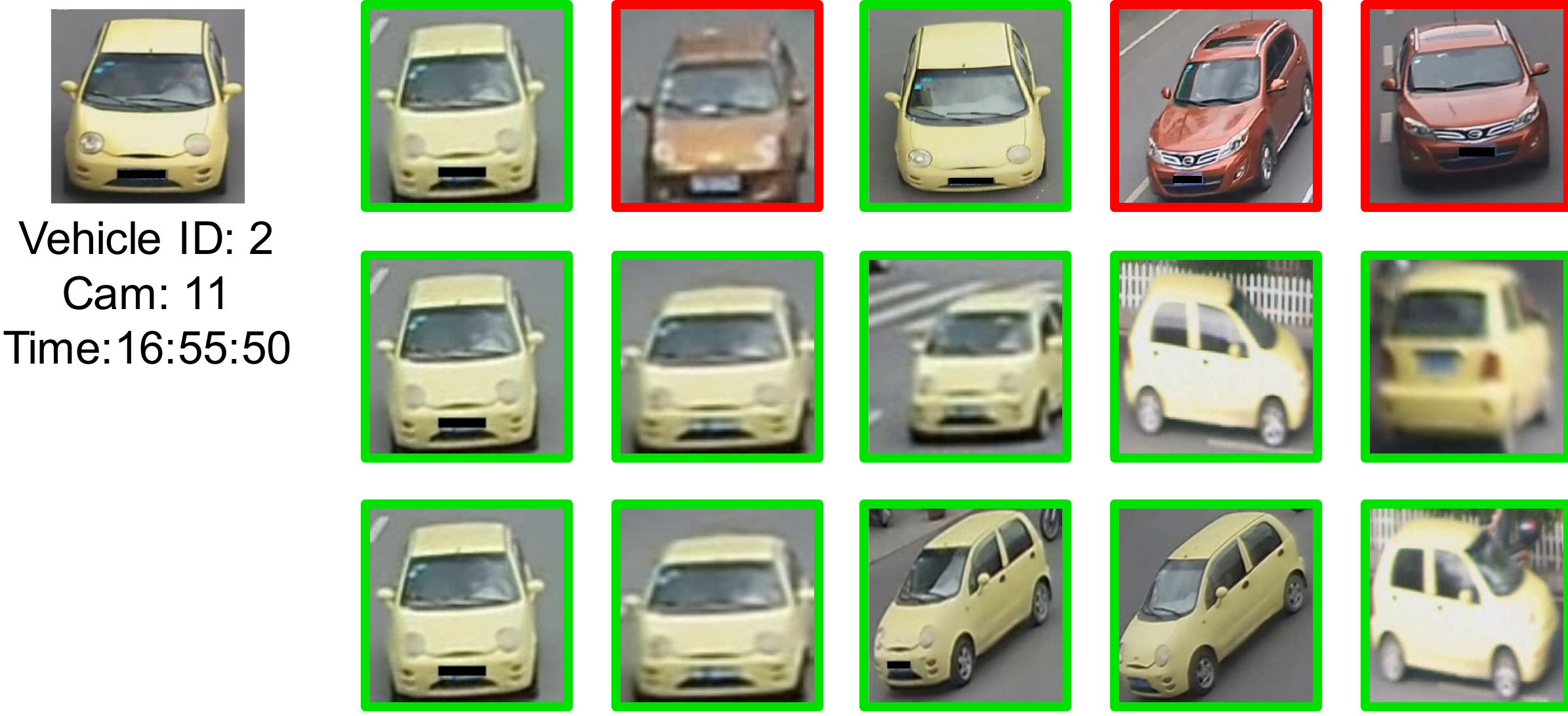}\\
   \\
   &\includegraphics[scale=0.27]{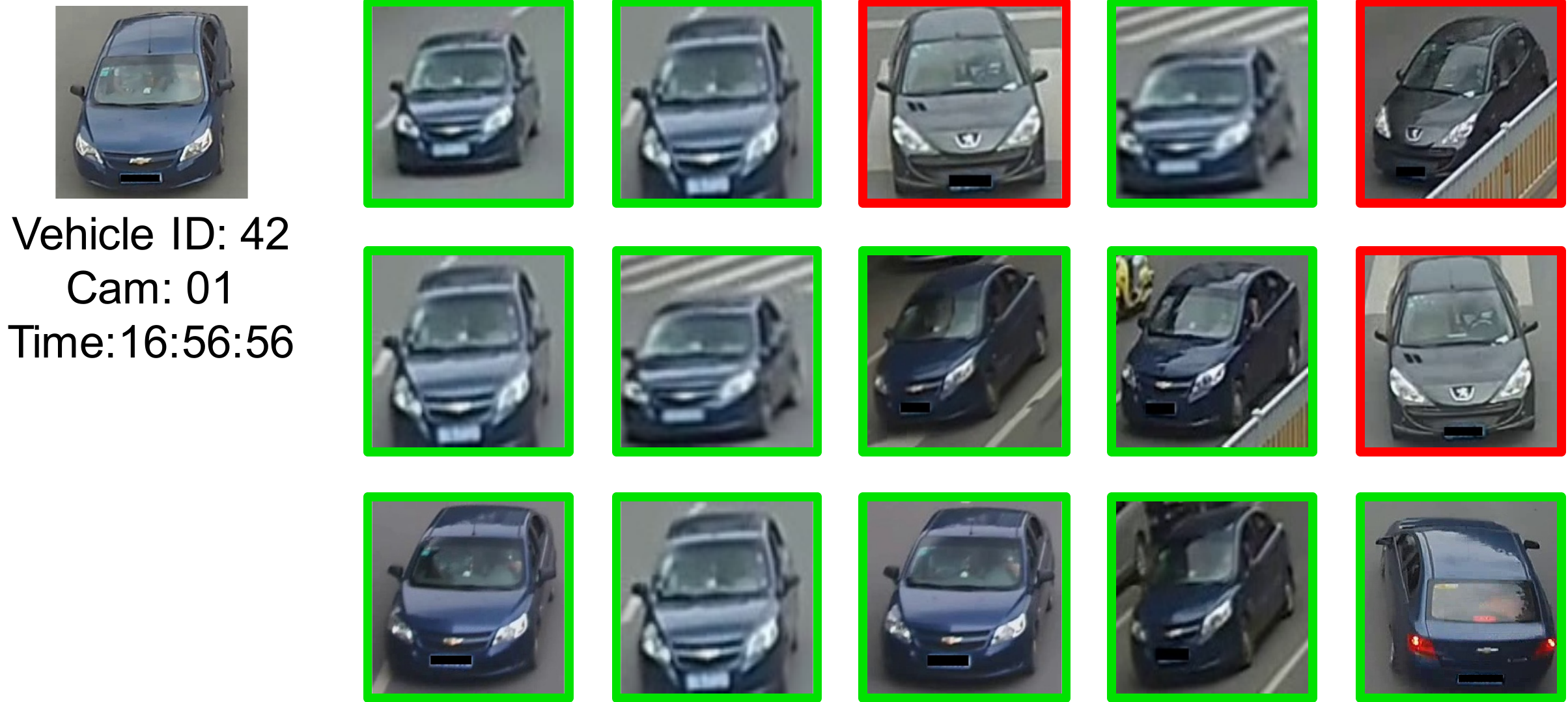}\\
\end{tabular}
\vspace{-12pt}
   \caption{Example vehicle re-identification results (top5) by our proposed approach. The true positive is in green box otherwise red. The three rows are results of Siamese-Visual, Siamese-CNN and Siamese-CNN+Path-LSTM.}
\label{fig:re-identification_results}
\end{figure}

The mAP, top-1 and top-5 accuracies of methods are listed in Tables \ref{tab:mAP} and \ref{tab:top}. Figure \ref{fig:CMC} shows the CMC curves of the compared methods. Example vehicle re-identification results by our approach are shown in Figure \ref{fig:re-identification_results}.

Our proposed two-stage approach, \emph{Siamese-CNN+Path-LSTM}, outperforms state-of-the-art methods \cite{liu2016large, liu2016deep} and all compared baselines, which demonstrates the effectiveness of our overall framework and individual components. 
Compared with \emph{Siamese-CNN}, which only takes pairwise visual and spatio-temporal information into account, our final approach has a gain of $4\%$ in terms of mAP and top-1 accuracy. Such a performance increase shows that the Path-LSTM with visual-spatio-temporal path proposal does provide vital priors for robustly estimating the vehicle similarities. Compared with \emph{Path-LSTM only}, which only calculates path-validness scores with the proposed visual-spatio-temporal path, our final approach also has a $4\%$ increase in terms of mAP and top-1 accuracy. This is because to generate the candidate path, our proposed chain MRF model always tries to discover the most feasible visual-spatio-temporal path. The visual-spatio-temporal state changes along the paths might sometimes be subtle and difficult to be captured by only pairwise differences of states at neighboring cameras. The obvious state difference between the query pair can sometimes be more easily captured by the \emph{Siamese-CNN}. Therefore, the \emph{Path-LSTM} acts as a strong prior for regularizing the \emph{Siamese-CNN} results and their combination shows the best retrieval performance.

Compared with the \emph{Chain MRF model}, the \emph{Path-LSTM} has a $10\%$ mAP gain and an increase of $25\%$ top-1 accuracy. Such results demonstrate that the empirical average is not a robust path-validness indicator of the candidate paths. Our trained \emph{Path-LSTM} is able to capture more subtle state changes on the candidate path for estimating correct path-validness scores. Compared with \emph{Siamese-Visual}, \emph{Siamese-CNN} has significant gains on mAP ($\sim$ $25\%$) and top-1 accuracy ($\sim$ $40\%$). It demonstrates that, unlike person re-identification, the spatio-temporal information is vital for vehicle re-identification, where the visual differences between different vehicles might be subtle for vehicles with the same color. Compared with \emph{Siamese-Visual+STR}, which adopts the spatio-temporal relation score in \cite{liu2016deep}, our \emph{Siamese-CNN} achieves more accurate retrieval performance. Our deep neural network is able to capture more complex spatio-temporal relations between query pairs.

The Path-LSTM shows strong capability on regularizing the retrieval results with candidate visual-spatio-temporal paths. The effectiveness of Path-LSTM relies on the correctness of the candidate paths. If the candidate path does correspond to the actual path, the Path-LSTM might have negative impact on the final similarity score. For all pairs of queries that have the same vehicle identities, we obtain their ground-truth visual-spatio-temporal paths and compare them with our proposed ones. The averaged Jaccard Similarity is calculated. Our proposed chain MRF model with deeply learned potential function achieves an AJS of 96.39\%.

We also test replacing ResNet50 with VGG16 in our pipeline. Our proposed overall framework (Siamese-VGG16+ PathLSTM-VGG16) and individual components (Path-LSTM-VGG16) outperform our VGG16 baseline (Siamese-CNN-VGG16).

\subsection{Time Complexity Analysis}
\label{ssec:timeca}
In the worst case, all pairwise potentials need to be calculated, resulting a complexity of $O(MK^2)$, where $M$ is the number of edges (connecting neighboring cameras) in the camera network, and $K$ is the average number of states at each camera. After that, the time complexity of dynamic programming for all path proposals is also $O(MK^2)$, which utilizes the technique described in Section \ref{cmrf} to avoid redundant computation. Amortized over the total number of $Q$ query pairs, each query pair has an averaged time complexity of $O(MK^2/Q)$. In practice, for testing, pairwise scores are calculated between 19.4 million query pairs (11579 galleries and 1678 queries). Because of our systematic way of avoiding redundant computation, each query pair only requires 0.016s on average.

\section{Conclusions}
In this paper, we proposed a two-stage framework for vehicle re-identification with both visual and spatio-temporal information. Existing methods ignored or used limited spatio-temporal information for regularizing the re-identification results. Our proposed approach incorporates important visual-spatial-temporal path information for regularization. A chain MRF model with deeply learned pairwise potential function is adopted to generate visual-spatio-temporal path proposals. Such candidate path proposals are evaluated by a Siamese-CNN+Path-LSTM to obtain similarity scores between pairs of queries. The proposed approach outperforms state-of-the-arts methods on the VeRi-776 dataset. Extensive component analysis of our framework demonstrates the effectiveness of our overall framework and individual components.

\textbf{Acknowledgments. }This work is supported in part by SenseTime Group Limited, in part by the General Research Fund through the Research Grants Council of Hong Kong under Grants CUHK14213616, CUHK14206114, CUHK14205615, CUHK419412, CUHK14203015, CUHK14239816, CUHK14207814, in part by the Hong Kong Innovation and Technology Support Programme Grant ITS/121/15FX, and in part by the China Postdoctoral Science Foundation under Grant 2014M552339.

{\small
\bibliographystyle{ieee}
\bibliography{egbib}

\begin{thebibliography}{10}\itemsep=-1pt

\bibitem{ahmed2015improved}
E.~Ahmed, M.~Jones, and T.~K. Marks.
\newblock An improved deep learning architecture for person re-identification.
\newblock In {\em Proceedings of the IEEE Conference on Computer Vision and
  Pattern Recognition}, pages 3908--3916, 2015.

\bibitem{bahdanau2014neural}
D.~Bahdanau, K.~Cho, and Y.~Bengio.
\newblock Neural machine translation by jointly learning to align and
  translate.
\newblock {\em arXiv preprint arXiv:1409.0473}, 2014.

\bibitem{cheng2016person}
D.~Cheng, Y.~Gong, S.~Zhou, J.~Wang, and N.~Zheng.
\newblock Person re-identification by multi-channel parts-based cnn with
  improved triplet loss function.
\newblock In {\em Proceedings of the IEEE Conference on Computer Vision and
  Pattern Recognition}, pages 1335--1344, 2016.

\bibitem{cho2014properties}
K.~Cho, B.~Van~Merri{\"e}nboer, D.~Bahdanau, and Y.~Bengio.
\newblock On the properties of neural machine translation: Encoder-decoder
  approaches.
\newblock {\em arXiv preprint arXiv:1409.1259}, 2014.

\bibitem{chu2015multi}
X.~Chu, W.~Ouyang, W.~Yang, and X.~Wang.
\newblock Multi-task recurrent neural network for immediacy prediction.
\newblock In {\em Proceedings of the IEEE International Conference on Computer
  Vision}, pages 3352--3360, 2015.

\bibitem{cormen2009introduction}
T.~H. Cormen.
\newblock {\em Introduction to algorithms}.
\newblock MIT press, 2009.

\bibitem{cucchiara2007multi}
R.~Cucchiara, A.~Prati, and R.~Vezzani.
\newblock A multi-camera vision system for fall detection and alarm generation.
\newblock {\em Expert Systems}, 24(5):334--345, 2007.

\bibitem{ding2015deep}
S.~Ding, L.~Lin, G.~Wang, and H.~Chao.
\newblock Deep feature learning with relative distance comparison for person
  re-identification.
\newblock {\em Pattern Recognition}, 48(10):2993--3003, 2015.

\bibitem{ellis2003learning}
T.~Ellis, D.~Makris, and J.~Black.
\newblock Learning a multi-camera topology.
\newblock In {\em Joint IEEE Workshop on Visual Surveillance and Performance
  Evaluation of Tracking and Surveillance (VS-PETS)}, pages 165--171, 2003.

\bibitem{feris2012large}
R.~S. Feris, B.~Siddiquie, J.~Petterson, Y.~Zhai, A.~Datta, L.~M. Brown, and
  S.~Pankanti.
\newblock Large-scale vehicle detection, indexing, and search in urban
  surveillance videos.
\newblock {\em IEEE Transactions on Multimedia}, 14(1):28--42, 2012.

\bibitem{graves2014towards}
A.~Graves and N.~Jaitly.
\newblock Towards end-to-end speech recognition with recurrent neural networks.
\newblock In {\em ICML}, volume~14, pages 1764--1772, 2014.

\bibitem{hamdoun2008person}
O.~Hamdoun, F.~Moutarde, B.~Stanciulescu, and B.~Steux.
\newblock Person re-identification in multi-camera system by signature based on
  interest point descriptors collected on short video sequences.
\newblock In {\em Distributed Smart Cameras, 2008. ICDSC 2008. Second ACM/IEEE
  International Conference on}, pages 1--6. IEEE, 2008.

\bibitem{he2016deep}
K.~He, X.~Zhang, S.~Ren, and J.~Sun.
\newblock Deep residual learning for image recognition.
\newblock In {\em Proceedings of the IEEE Conference on Computer Vision and
  Pattern Recognition}, pages 770--778, 2016.

\bibitem{hochreiter1997long}
S.~Hochreiter and J.~Schmidhuber.
\newblock Long short-term memory.
\newblock {\em Neural computation}, 9(8):1735--1780, 1997.

\bibitem{jaccard1901etude}
P.~Jaccard.
\newblock {\em Etude comparative de la distribution florale dans une portion
  des Alpes et du Jura}.
\newblock Impr. Corbaz, 1901.

\bibitem{javed2008modeling}
O.~Javed, K.~Shafique, Z.~Rasheed, and M.~Shah.
\newblock Modeling inter-camera space--time and appearance relationships for
  tracking across non-overlapping views.
\newblock {\em Computer Vision and Image Understanding}, 109(2):146--162, 2008.

\bibitem{kang2017tpn}
K.~Kang, H.~Li, T.~Xiao, W.~Ouyang, J.~Yan, X.~Liu, and X.~Wang.
\newblock Object detection in videos with tubelet proposal networks.
\newblock In {\em CVPR}, 2017.

\bibitem{karpathy2015deep}
A.~Karpathy and L.~Fei-Fei.
\newblock Deep visual-semantic alignments for generating image descriptions.
\newblock In {\em Proceedings of the IEEE Conference on Computer Vision and
  Pattern Recognition}, pages 3128--3137, 2015.

\bibitem{kettnaker1999bayesian}
V.~Kettnaker and R.~Zabih.
\newblock Bayesian multi-camera surveillance.
\newblock In {\em Computer Vision and Pattern Recognition, 1999. IEEE Computer
  Society Conference on.}, volume~2, pages 253--259. IEEE, 1999.

\bibitem{kingma2014adam}
D.~Kingma and J.~Ba.
\newblock Adam: A method for stochastic optimization.
\newblock {\em arXiv preprint arXiv:1412.6980}, 2014.

\bibitem{koestinger2012large}
M.~Koestinger, M.~Hirzer, P.~Wohlhart, P.~M. Roth, and H.~Bischof.
\newblock Large scale metric learning from equivalence constraints.
\newblock In {\em Computer Vision and Pattern Recognition (CVPR), 2012 IEEE
  Conference on}, pages 2288--2295. IEEE, 2012.

\bibitem{krizhevsky2012imagenet}
A.~Krizhevsky, I.~Sutskever, and G.~E. Hinton.
\newblock Imagenet classification with deep convolutional neural networks.
\newblock In {\em Advances in neural information processing systems}, pages
  1097--1105, 2012.

\bibitem{DBLP:journals/corr/LiLOW17}
H.~Li, Y.~Liu, W.~Ouyang, and X.~Wang.
\newblock Zoom out-and-in network with recursive training for object proposal.
\newblock {\em CoRR}, abs/1702.05711, 2017.

\bibitem{li2017person}
S.~Li, T.~Xiao, H.~Li, B.~Zhou, D.~Yue, and X.~Wang.
\newblock Person search with natural language description.
\newblock In {\em CVPR}, 2017.

\bibitem{livip}
Y.~Li, W.~Ouyang, X.~Wang, and X.~Tang.
\newblock Vip-cnn: Visual phrase guided convolutional neural network.
\newblock In {\em CVPR}, 2017.

\bibitem{liu2016deepfine}
H.~Liu, Y.~Tian, Y.~Yang, L.~Pang, and T.~Huang.
\newblock Deep relative distance learning: Tell the difference between similar
  vehicles.
\newblock In {\em Proceedings of the IEEE Conference on Computer Vision and
  Pattern Recognition}, pages 2167--2175, 2016.

\bibitem{liu2016large}
X.~Liu, W.~Liu, H.~Ma, and H.~Fu.
\newblock Large-scale vehicle re-identification in urban surveillance videos.
\newblock In {\em Multimedia and Expo (ICME), 2016 IEEE International
  Conference on}, pages 1--6. IEEE, 2016.

\bibitem{liu2016deep}
X.~Liu, W.~Liu, T.~Mei, and H.~Ma.
\newblock A deep learning-based approach to progressive vehicle
  re-identification for urban surveillance.
\newblock In {\em European Conference on Computer Vision}, pages 869--884.
  Springer, 2016.

\bibitem{loy2009multi}
C.~C. Loy, T.~Xiang, and S.~Gong.
\newblock Multi-camera activity correlation analysis.
\newblock In {\em Computer Vision and Pattern Recognition, 2009. CVPR 2009.
  IEEE Conference on}, pages 1988--1995. IEEE, 2009.

\bibitem{mahendran2015car}
S.~Mahendran and R.~Vidal.
\newblock Car segmentation and pose estimation using 3d object models.
\newblock {\em arXiv preprint arXiv:1512.06790}, 2015.

\bibitem{matei2011vehicle}
B.~C. Matei, H.~S. Sawhney, and S.~Samarasekera.
\newblock Vehicle tracking across nonoverlapping cameras using joint kinematic
  and appearance features.
\newblock In {\em Computer Vision and Pattern Recognition (CVPR), 2011 IEEE
  Conference on}, pages 3465--3472. IEEE, 2011.

\bibitem{mcfee2010metric}
B.~McFee and G.~R. Lanckriet.
\newblock Metric learning to rank.
\newblock In {\em Proceedings of the 27th International Conference on Machine
  Learning (ICML-10)}, pages 775--782, 2010.

\bibitem{neumann2002spatio}
J.~Neumann and Y.~Aloimonos.
\newblock Spatio-temporal stereo using multi-resolution subdivision surfaces.
\newblock {\em International Journal of Computer Vision}, 47(1-3):181--193,
  2002.

\bibitem{paisitkriangkrai2015learning}
S.~Paisitkriangkrai, C.~Shen, and A.~van~den Hengel.
\newblock Learning to rank in person re-identification with metric ensembles.
\newblock In {\em Proceedings of the IEEE Conference on Computer Vision and
  Pattern Recognition}, pages 1846--1855, 2015.

\bibitem{sochor2016boxcars}
J.~Sochor, A.~Herout, and J.~Havel.
\newblock Boxcars: 3d boxes as cnn input for improved fine-grained vehicle
  recognition.
\newblock In {\em Proceedings of the IEEE Conference on Computer Vision and
  Pattern Recognition}, pages 3006--3015, 2016.

\bibitem{szegedy2015going}
C.~Szegedy, W.~Liu, Y.~Jia, P.~Sermanet, S.~Reed, D.~Anguelov, D.~Erhan,
  V.~Vanhoucke, and A.~Rabinovich.
\newblock Going deeper with convolutions.
\newblock In {\em Proceedings of the IEEE Conference on Computer Vision and
  Pattern Recognition}, pages 1--9, 2015.

\bibitem{wang2007shape}
X.~Wang, G.~Doretto, T.~Sebastian, J.~Rittscher, and P.~Tu.
\newblock Shape and appearance context modeling.
\newblock In {\em Computer Vision, 2007. ICCV 2007. IEEE 11th International
  Conference on}, pages 1--8. IEEE, 2007.

\bibitem{weinberger2009distance}
K.~Q. Weinberger and L.~K. Saul.
\newblock Distance metric learning for large margin nearest neighbor
  classification.
\newblock {\em Journal of Machine Learning Research}, 10(Feb):207--244, 2009.

\bibitem{wu2016model}
F.~Wu, S.~Li, T.~Zhao, and K.~N. Ngan.
\newblock Model-based face reconstruction using sift flow registration and
  spherical harmonics.
\newblock In {\em Pattern Recognition (ICPR), 2016 23rd International
  Conference on}, pages 1774--1779. IEEE, 2016.

\bibitem{xiao2016learning}
T.~Xiao, H.~Li, W.~Ouyang, and X.~Wang.
\newblock Learning deep feature representations with domain guided dropout for
  person re-identification.
\newblock In {\em Proceedings of the IEEE Conference on Computer Vision and
  Pattern Recognition}, pages 1249--1258, 2016.

\bibitem{xiaoli2017joint}
T.~Xiao, S.~Li, B.~Wang, L.~Lin, and X.~Wang.
\newblock Joint detection and identification feature learning for person
  search.
\newblock In {\em CVPR}, 2017.

\bibitem{yang2015large}
L.~Yang, P.~Luo, C.~Change~Loy, and X.~Tang.
\newblock A large-scale car dataset for fine-grained categorization and
  verification.
\newblock In {\em Proceedings of the IEEE Conference on Computer Vision and
  Pattern Recognition}, pages 3973--3981, 2015.

\bibitem{yi2014deep}
D.~Yi, Z.~Lei, S.~Liao, and S.~Z. Li.
\newblock Deep metric learning for person re-identification.
\newblock In {\em Pattern Recognition (ICPR), 2014 22nd International
  Conference on}, pages 34--39. IEEE, 2014.

\bibitem{yue2015beyond}
J.~Yue-Hei~Ng, M.~Hausknecht, S.~Vijayanarasimhan, O.~Vinyals, R.~Monga, and
  G.~Toderici.
\newblock Beyond short snippets: Deep networks for video classification.
\newblock In {\em Proceedings of the IEEE conference on computer vision and
  pattern recognition}, pages 4694--4702, 2015.

\bibitem{zapletal2016vehicle}
D.~Zapletal and A.~Herout.
\newblock Vehicle re-identification for automatic video traffic surveillance.
\newblock In {\em Proceedings of the IEEE Conference on Computer Vision and
  Pattern Recognition Workshops}, pages 25--31, 2016.

\end{thebibliography}
}

\end{document}